\documentclass[onecolumn,english]{article}
\usepackage[T1]{fontenc}
\usepackage[utf8]{luainputenc}
\usepackage{array}
\usepackage{amsmath}
\usepackage{graphicx}
\usepackage{babel}
\usepackage{natbib}

\usepackage[accepted]{whi2016} 

\icmltitlerunning{Explainable Restricted Boltzmann Machines for Collaborative Filtering}
\begin{document}

\twocolumn[
\icmltitle{Explainable Restricted Boltzmann Machines for Collaborative Filtering}
\icmlauthor{Behnoush Abdollahi}{b.abdollahi@louisville.edu}
\icmladdress{Dept. of Computer Engineering \& Computer
	Science, Knowledge Discovery and Web Mining Lab, University of Louisville,
	Louisville, KY 40222, USA}
\icmlauthor{Olfa Nasraoui}{olfa.nasraoui@louisville.edu}
\icmladdress{Dept. of Computer Engineering \& Computer
	Science, Knowledge Discovery and Web Mining Lab, University of Louisville,
	Louisville, KY 40222, USA}
% You may provide any keywords that you 
% find helpful for describing your paper; these are used to populate 
% the "keywords" metadata in the PDF but will not be shown in the document
%\icmlkeywords{boring formatting information, machine learning, ICML}

\vskip 0.3in
]
\begin{abstract}
Most accurate recommender systems are black-box models, hiding the reasoning behind their recommendations. Yet explanations have been shown to increase the user's trust in the system in addition to providing other benefits such as scrutability, meaning the ability to verify the validity of recommendations. This gap between accuracy and
transparency or explainability has generated an interest in
automated explanation generation methods. Restricted Boltzmann Machines (RBM) are accurate models for CF that also lack interpretability.
In this paper, we focus on RBM based collaborative filtering recommendations, and further assume the absence of
any additional data source, such as item content or user attributes. We thus propose a new Explainable RBM technique that computes the top-$n$ recommendation list from
items that are explainable. Experimental results show that our method is effective
in generating accurate and explainable recommendations.
\end{abstract}
\vspace{-10pt}
\section{Introduction}
Explanations for recommendations can have several benefits, such as:
helping the user make a good decision (effectiveness), helping the
user make a faster decision (efficiency), and revealing the reasoning
behind the system's recommendation (transparency) \cite{tintarev2011designing,Zanker:2012:IKE:2365952.2366011}.
As a result, users are more likely to follow the recommendation and
use the system in better ways \cite{tintarev2007survey,herlocker2000explaining}.
For instance, the Netflix recommender system justifies its movie suggestions
by listing similar movies, obtained from the user's social network.
Amazon's recommender system shows similar items to the ones that the
user (or other similar users) have bought or viewed, when recommending
a new item using neighborhood based Collaborative Filtering (CF).

CF approches provide recommendations to users based on their \emph{collective}
recorded interests on items, typically relying on the similarity between
users or items, giving rise to neighborhood-based CF approaches, which
can be user-based or item-based. Neighborhood-based CF methods are
white-box approaches that can be explained based on the ratings of similar users or items. 

Most accurate recommender systems are model-based methods that are
black-boxes. Among model-based approaches are Restricted Boltzmann
Machines (RBM) \cite{hinton2010practical} that can assign a low dimensional set of features to
items in a latent space. The newly obtained set of features capture the user's
interests and different items groups; however, it is very difficult
to interpret these automatically learned features. Therefore, justification
of the recommendation or the reasoning behind the recommended item
in these models is not clear. RBM approaches have recently proved to be powerful for designing
deep learning techniques to learn and predict patterns in large datasets
because they can provide very accurate results \cite{hinton2006reducing}. However, they suffer from the lack of interpretation of the results, especially for
recommender systems.
Lack of explanations can result in users not trusting the suggestions
made by the recommender system. Therefore, the only way for the user
to assess the quality of a recommendation is by following it. This,
however, is contrary to one of the goals of a recommendation system,
which is reducing the time that users spend on exploring items. It
would be very desirable and beneficial to design recommender systems
that can give accurate suggestions, which, at the same time, facilitate
conveying the reasoning behind the recommendations to the user. A
main challenge in creating a recommender system is to choose an interpretable
technique with moderate prediction accuracy or a more accurate technique,
such RBM, which does not give explainable recommendations.
\subsection{Problem Statement}
Our research question is: can we design an RBM model for a CF recommender
engine that suggests items that are explainable, while recommendations
remain accurate? Our current scope is limited to CF recommendations
where no additional source of data is used in explanations, and where
explanations for recommended items can be generated from the ratings
given to these items, by the active user's neighbors only (user-based
neighbor style explanation), as shown in Figure \ref{fig:example-explanation}. 
\begin{table*}
	\protect\caption{\label{tab:example-1-1}Top-3 recommendations and explanations for
		a test user.}
	\begin{tabular}{|c|c|>{\centering}p{0.6\textwidth}|}
		\hline 
		\multicolumn{1}{|c|}{{\tiny{}top-3 test use ratings}} & \multicolumn{2}{c|}{{\tiny{}Explainable RBM}}\tabularnewline
		\hline 
		{\tiny{}movie} & \centering{}{\tiny{}recommended movie} & \centering{}{\tiny{}explanation (ratings out of 5)}\tabularnewline
		\hline 
		{\tiny{}Annie Hall (Comedy)} & \centering{}{\tiny{}Miller's Crossing (Thriller)} & \centering{}{\tiny{}8 out of 10 people with similar interests to you
			have rated this movie 4 and higher.}\tabularnewline
		\hline 
		{\tiny{}Carrie (Drama)} & {\tiny{}Sinin' in the Rain (Comedy)} & {\tiny{}9 out of 10 people with similar interests to you have rated
			this movie 3 and higher.}\tabularnewline
		\hline 
		{\tiny{}Jaws (Thriller)} & {\tiny{}Psycho (Horror)} & {\tiny{}4 out of 10 people with similar interests to you have rated
			this movie 5.}\tabularnewline
		\hline 
	\end{tabular}
\end{table*}
\section{Related Work}
\subsection{Explaining CF}
\begin{figure}
	\begin{centering}
		\begin{tabular}{cc}
			\includegraphics[width=0.8\columnwidth]{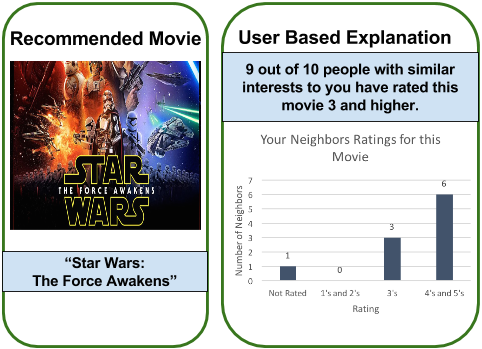}\tabularnewline
		\end{tabular}
		\par\end{centering}	
	\protect\caption{\label{fig:example-explanation}An example of a user-based neighbor
		style explanation for a recommended item.}
\end{figure}

There are different ways of classifying explanation styles. Generally
explanations can be user-based neighbor-style (Figure \ref{fig:example-explanation}),
item-based neighbor style (also known as influence-style), and keyword-style 
\cite{bilgic2005explaining}. A user-based neighbor-style explanation
is based on similar users, and generally is used when the CF method
is also a user-based neighborhood style method. An item-based neighbor-style
explanation is generally used in item-based CF methods by presenting
the items that had the highest impact on the recommender system's
decision. A keyword-style explanation is based on the items' features
or users' demographic data available as content data and is mostly
used in content based recommender systems \cite{bilgic2005explaining}.

In all styles, the explanation \emph{may}, or \emph{may not} reflect
the underlying algorithm used by the recommender system. Also, data
sources employed in the recommendation task, may be different from
the data sources used in generating the explanation \cite{vig2009tagsplanations,herlocker2000explaining,symeonidis2008justified,bilgic2005explaining,billsus1999personal}.
Hence, the explanation generation is a separate module from the recommender
system. However, performing the recommendation task based on the items'
explainability (thus \emph{integrating} recommendation and explanation)
may improve the transparent suggestion of interpretable items to the
user, while enjoying the powerful prediction of a model-based CF approach.

Zhang et al. \cite{zhang2014explicit} proposed a model-based CF to
generate explainable recommendations based on item features and sentiment
analysis of user reviews, as data sources, in addition to the ratings
data. That said, their approach is similar to our proposed method in that the recommender model suggests
highly explainable items as recommendations and in that the recommendation
and explanation engines are \emph{not} separate. They further expanded
their feature-level explanation by considering different forms such
as word clouds \cite{zhang2015incorporating}. Also, \cite{abdollahi2016explainable} proposed explainable MF (EMF) method for explainable CF. In contrast to Zhang et al. \cite{zhang2014explicit}, EMF approach
does \emph{not} require any additional data such as reviews for explanation generation. Herlocker et al. \cite{herlocker2000explaining}
performed a detailed study on 21 different styles of explanation generation
for neighbor-based CF methods, including content-based explanations
that present a list of features from the recommended item to the user.
Symeonidis et al. \cite{symeonidis2008justified} also proposed an
explanation approach based on the content features of the items. Their recommender system and the explanation approach are both content-based but they used different algorithms. Bilgic and Mooney \cite{bilgic2005explaining}
proposed three forms of explanations: keyword style, neighbor style
and influence style, as separate forms of explanation approaches from
their recommender system, a hybrid approach, called LIBRA, that recommends
books \cite{mooney2000content}. Billsus and Pazzani \cite{billsus1999personal}
presented a keyword style and influence style explanation approach
for their news recommendation system. The system generates explanations and adapts
its recommendation to the user's interests based on the user's preferences
and interests. 
In all the reviewed existing approaches except \cite{herlocker2000explaining}, the explanations have to resort to external data such as item content or reviews, while useful when available, are outside the scope of the
work in this paper that primarily aims at explainability of pure CF without resorting to external data.

%\begin{figure}
%\begin{centering}
%\begin{tabular}{cc}
%\includegraphics[width=0.38\columnwidth]{../../abstract/images/exam_tab} & \includegraphics[width=0.58\columnwidth]{../../abstract/images/example}\tabularnewline
%\end{tabular}
%\par\end{centering}
%
%\protect\caption{\label{fig:example-explanation}An example of a user-based neighbor
%style explanation presented using two formats.}
%\end{figure}
%
%
%\begin{figure}
%\begin{centering}
%\begin{tabular}{cc}
%\includegraphics[scale=0.35]{../../abstract/images/Classical} & \includegraphics[scale=0.35]{../../abstract/images/ourapproach}\tabularnewline
%(a) & (b)\tabularnewline
%\end{tabular}
%\par\end{centering}
%\protect\caption{\label{fig:frames} Conventional framework for recommendation and
%explanation generation (left) and the proposed framework for explainable
%recommendations (right).}
%\end{figure}
\subsection{RBM for CF}
RBM is a two layer stochastic neural network consisting of visible
and hidden units. Each visible unit is connected to all the hidden
units in an undirected form. No visible/hidden unit is connected to
any other visible/hidden unit. The stochastic, binary visible units
encode user preferences on the items from the training data, therefore the state of every visible unit is known. Hidden units are also stochastic,
binary variables that capture the latent features. A probability $p(\textbf{v},\textbf{h})$
is assigned to each pair of a hidden and a visible vector:\begin{equation}
p(\textbf{v},\textbf{h})=\frac{e^{-E(\textbf{v},\textbf{h})}}{Z}
\end{equation}
where $E$ is the energy of the system and $Z$ is a normalizing
factor, as defined in \cite{Hinton:2002:TPE:639729.639730}.To train
for the weights, a Contrastive Divergence method was proposed by Hinton \cite{Hinton:2002:TPE:639729.639730}. Salakhutdinov
et al. \cite{Salakhutdinov:2007:RBM:1273496.1273596}, proposed an
RBM framework for CF. Their model assumes one RBM for each user and
takes only rated items into consideration when learning the weights.
They presented the results of their approach on the Netflix data and
showed that their technique was more accurate than Netflix's own system.
The focus of this RBM approach was only on the accuracy and predicting error
and not explanation generation.
\section{Proposed Methods}
In this section, we present our explainable RBM framework. First, we
present our approach for measuring explainability of each item for
each user using explicit rating data. Next, we propose the explainable
RBM framework.
\begin{figure}
	\begin{centering}
		\includegraphics[width=0.7\columnwidth]{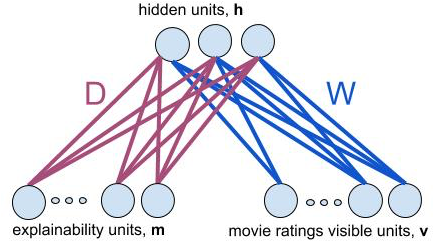}
		\par\end{centering}
	\protect\caption{\label{fig:Conditional-RBM}Conditional RBM for explainabilty.}
\end{figure}
\section{Explainability}
Explainability can be formulated based on the rating distribution
within the active user's neighborhood. The main idea is that if many
neighbors have rated the recommended item, then this could provide
a basis upon which to explain the recommendations, using neighborhood
style explanation mechanisms. For user-based neighbor-style explanations,
such as the ones shown in Figure \ref{fig:example-explanation}, we
can therefore define the \emph{Explainability Score} of item $i$
for user $u$ as follows:
\begin{equation}
Expl.\,Score(u,i)=\frac{\sum_{x\in N_{k}(u)}r_{x,i}}{\sum_{N_{k}(u)}R_{max}}\label{eq:xp_ratio}
\end{equation}
where $N_{k}(u)$ is the set of user $u$'s $k$ neighbors, $r_{x,i}$
is the rating of $x$ on item $i$ and $R_{max}$ is the maximum rating
value of $N_{k}(u)$ on $i$. Neighbors are determined based on the cosine similarity. Without
loss of information, $r_{x,i}$ is considered as $0$ for missing
ratings, indicating that user $x$ does not contribute to the user-based
neighbor-style explanation of item $i$ for user $u$. Given this
definition, it is obvious that Explainability Score is between zero
and one. Item $i$ is explainable for user $u$ only if its explainability
score is larger than zero. When no explanation can be made, the explainability
ratio would be zero. 
%Based on the explainability ratio, we define
%the\emph{ Explainability Matrix}, $M$, between user-item pairs as
%follows:
%
%\begin{equation}
%M_{ui}=\begin{cases}
%\begin{array}{cc}
%1 & \:if\,Expl.\,Ratio(u,i)>\theta\\
%0 & otherwise
%\end{array}\end{cases}\label{eq:weight_mat}
%\end{equation}
%where $\theta$ denotes an optional threshold above which we accept
%item $i$ to be explainable for user $u$. $M_{ui}$ thus measures
%the explainability of item $i$ for user $u$.
\section{Conditional RBM}
The conditional RBM model takes explainability into account with an additional visible layer, $\textbf{m}$, with $n$ nodes, where $n$ is the number of items. Each node has a value between $0$ and $1$, indicating the explainability score of the relative item to the current user in the iteration, calculated as explained in Section $4$. The idea is to define a joint
distribution over $(\textbf{v},\textbf{h})$, conditional on the explainability scores, $\textbf{m}$. Figure \ref{fig:Conditional-RBM} presents the
conditional RBM model with explainability.
Based on \cite{hinton2010practical}, the $p(h_{j}=1|\textbf{v},\textbf{m})$, $p(v_{i}=1|\textbf{h})$, and $p(m_{i}=1|\textbf{h})$ are defined as:
\begin{equation}
p(h_{j}=1|\textbf{v},\textbf{m})=\sigma(a_{j}+\sum_{i=1}^{n}v{}_{i}W_{ij}+\sum_{i=1}^{n}m_{i}D_{ij})
\vspace{-10pt}
\end{equation}
\begin{equation}
p(v_{i}=1|\textbf{h})=\sigma(b_{i}+\sum_{j=1}^{f}h_{j}W_{ij})
\end{equation}
\begin{equation}
p(m_{i}=1|\textbf{h})=\sigma(c_{i}+\sum_{j=1}^{f}h_{j}D_{ij})
\end{equation}
where $a$, $b$, and $c$ are biases, $f$ and $n$ are the numbers of hidden and visible units, respectively, and $W$ and $D$ are the weights of the entire network. $\sigma(x)$ is the logistic function $\frac{1}{(1+e^{-x})}$.

To avoid computing a model, we follow an approximation to the gradient
of a different objective function called “Contrastive Divergence”
(CD) \cite{Hinton:2002:TPE:639729.639730}:
\begin{equation}
\text{\ensuremath{\Delta}}W_{ij}=\epsilon_{W}(<v_{i}h_{j}>_{data}-<v_{i}h_{j}>_{recom})
\end{equation}
where $W_{ij}$ is an element of a learned matrix that models the
effect of ratings on $h$. Learning $D$, which is the effect of explainability on $h$, using CD, is similar and takes the form: 
\begin{equation}
\text{\ensuremath{\Delta}}D_{ij}=\epsilon_{D}(<m_{i}h_{j}>_{data}-<m_{i}h_{j}>_{recom})
\end{equation}
\section{Experimental results}
\begin{figure}
	\begin{centering}
		\begin{tabular}{cc}
			\includegraphics[width=0.5\columnwidth]{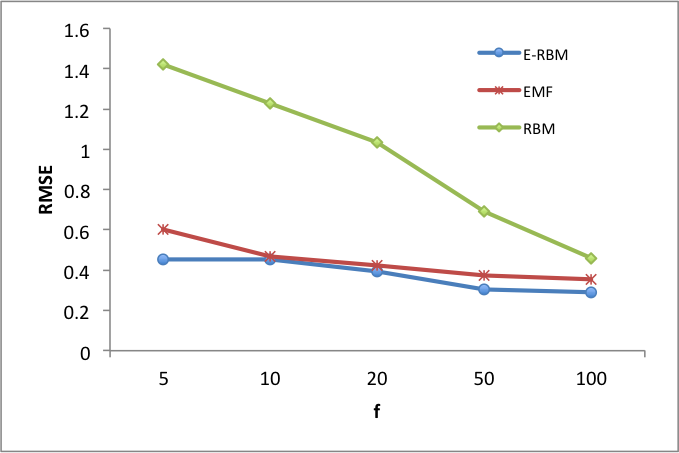} & \includegraphics[width=0.5\columnwidth]{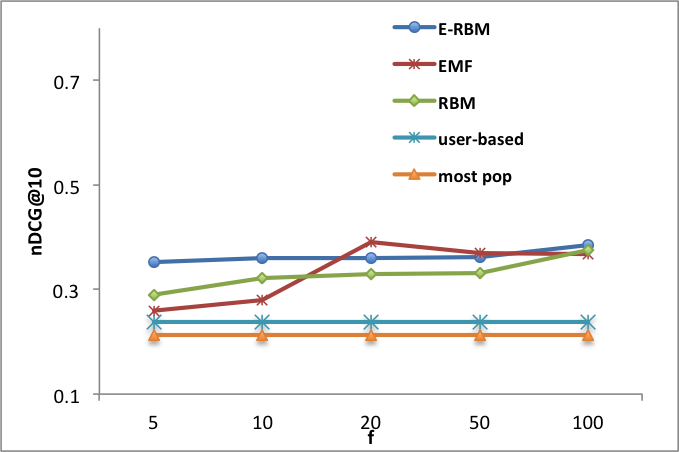}\tabularnewline
			\includegraphics[width=0.5\columnwidth]{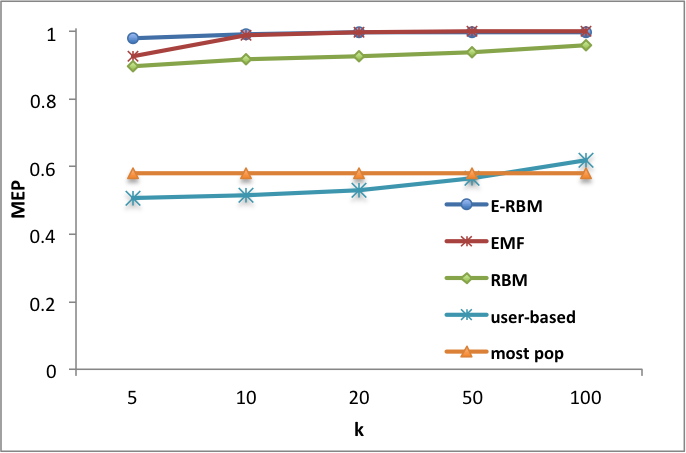} & \includegraphics[width=0.5\columnwidth]{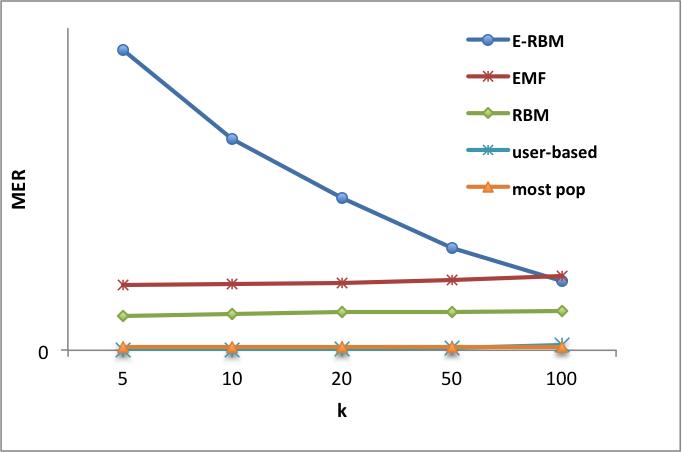}\tabularnewline
		\end{tabular}
		\par\end{centering}
	\vspace{-5pt}
	\protect\caption{\label{fig:MEPMER}RMSE, nDCG (top row) \emph{vs.} $f$, and MEP, MER (bottom row) \emph{vs.} $k$ (number of neighbors).}
	\vspace{-15pt}
\end{figure}

We tested our approach on the MovieLens \footnote{http://www.grouplens.org/node/12}
ratings data which consists of $100,000$ ratings, on a scale of $1$ to $5$, for $1700$ movies and $1000$ users. The data is split
into training and test sets such that $10\%$ of the latest ratings from each user are selected for the test set and the remaining are used in the training set. Ratings are normalized between $0$ and $1$, to be used as RBM input. We compare our results with RBM, Explainable Matrix Factorization \cite{abdollahi2016explainable}, user-based top-n recommendations \cite{herlocker1999algorithmic}, and non-personalized most popular items. Each Experiments is run $10$ times and the average results are reported. To assess the rating prediction accuracy, we used the Root Mean Squared Error (RMSE) metric: $RMSE=\sqrt{\sum_{(u,i)\in testset}(r_{ui}-\hat{r}_{ui})^{2}/|testset|}$. Note that RMSE can only be calculated for the prediction based methods and not the top-n recommendation techniques. To evaluate the top-n recommendation task, we used the nDCG@10 metric following \cite{herlocker2004evaluating}. The results for RMSE and nDCG, when varying the number of hidden units, $f$, is presented in Figure \ref{fig:MEPMER}, top row. It can be observed that Explainable RBM (E-RBM) outperforms other approaches when $f>20$ for RMSE and when $f<20$ for nDCG. To evaluate the explainability of the proposed approach, we used Mean Explainability Precision (MEP) and Mean Explainability Recall (MER) metrics \cite{abdollahi2016explainable} which are higher when the recommend items are explainable as per Eq (2). The results are shown in Figure \ref{fig:MEPMER}, bottom row. It can be observed that explainable RBM outperforms other approaches in terms of explainability. For a test user, the top-3 rated movies
and their genres by the user, in addition to the top-3 recommended movies and the explanations generated using the proposed method, are presented in Table \ref{tab:example-1-1}. Explanations can be presented to the users using the format shown in Figure \ref{fig:example-explanation}.

\section{Conclusion}
%We presented an RBM-based recommender
%system that can suggest items that are relevant and explainable in the latent space, without any major sacrifice
%in recommendation accuracy. Our scope, in this work, is
%limited to CF recommendations where no additional source
%of data is used in recommendations or in explanations, and
%where explanations for recommended items can be generated from the ratings given to these items, by the active
%user's neighbors only. Thus explainability can be directly
%formulated based on the rating distribution within the ac-
%tive user's neighborhood. This is because if many neighbors
%have rated the recommended item, then this provides a basis
%upon which one can explain the recommendations. We focused on user-based neighbor style explanations,
%however, a similar approach can be obtained for the item
%style explanations. In our current work, we have only incorporated the user-based neighbor style explanation and we
%did not use any external data source, but rather used only
%the ratings. This is important because using no external
%source is one of the main challenges for explainable.
We presented an explainable RBM approach for CF recommendations that achieves both accuracy and
interpretability by learning an RBM network that tries to estimate accurate user ratings while also
taking into account the explainability of an item to a user. Both rating prediction and explainability are
integrated within one learning goal allowing to learn a network model that prioritizes the recommendation
of items that are explainable.

\bibliography{sigproc,/Users/behnoush/Dropbox/abstract/sigproc}

\begin{thebibliography}{18}
\providecommand{\natexlab}[1]{#1}
\providecommand{\url}[1]{\texttt{#1}}
\expandafter\ifx\csname urlstyle\endcsname\relax
  \providecommand{\doi}[1]{doi: #1}\else
  \providecommand{\doi}{doi: \begingroup \urlstyle{rm}\Url}\fi

\bibitem[Abdollahi \& Nasraoui(2016)Abdollahi and
  Nasraoui]{abdollahi2016explainable}
Abdollahi, Behnoush and Nasraoui, Olfa.
\newblock Explainable matrix factorization for collaborative filtering.
\newblock In \emph{Proceedings of the 25th International Conference Companion
  on World Wide Web}, pp.\  5--6. International World Wide Web Conferences
  Steering Committee, 2016.

\bibitem[Bilgic \& Mooney(2005)Bilgic and Mooney]{bilgic2005explaining}
Bilgic, Mustafa and Mooney, Raymond~J.
\newblock Explaining recommendations: Satisfaction vs. promotion.
\newblock In \emph{Beyond Personalization Workshop, IUI}, volume~5, 2005.

\bibitem[Billsus \& Pazzani(1999)Billsus and Pazzani]{billsus1999personal}
Billsus, Daniel and Pazzani, Michael~J.
\newblock A personal news agent that talks, learns and explains.
\newblock In \emph{Proceedings of the third annual conference on Autonomous
  Agents}, pp.\  268--275. ACM, 1999.

\bibitem[Herlocker et~al.(1999)Herlocker, Konstan, Borchers, and
  Riedl]{herlocker1999algorithmic}
Herlocker, Jonathan~L, Konstan, Joseph~A, Borchers, Al, and Riedl, John.
\newblock An algorithmic framework for performing collaborative filtering.
\newblock In \emph{Proceedings of the 22nd annual international ACM SIGIR
  conference on Research and development in information retrieval}, pp.\
  230--237. ACM, 1999.

\bibitem[Herlocker et~al.(2000)Herlocker, Konstan, and
  Riedl]{herlocker2000explaining}
Herlocker, Jonathan~L, Konstan, Joseph~A, and Riedl, John.
\newblock Explaining collaborative filtering recommendations.
\newblock In \emph{Proceedings of the 2000 ACM conference on Computer supported
  cooperative work}, pp.\  241--250. ACM, 2000.

\bibitem[Herlocker et~al.(2004)Herlocker, Konstan, Terveen, and
  Riedl]{herlocker2004evaluating}
Herlocker, Jonathan~L, Konstan, Joseph~A, Terveen, Loren~G, and Riedl, John~T.
\newblock Evaluating collaborative filtering recommender systems.
\newblock \emph{ACM Transactions on Information Systems (TOIS)}, 22\penalty0
  (1):\penalty0 5--53, 2004.

\bibitem[Hinton(2010)]{hinton2010practical}
Hinton, Geoffrey.
\newblock A practical guide to training restricted boltzmann machines.
\newblock \emph{Momentum}, 9\penalty0 (1):\penalty0 926, 2010.

\bibitem[Hinton(2002)]{Hinton:2002:TPE:639729.639730}
Hinton, Geoffrey~E.
\newblock Training products of experts by minimizing contrastive divergence.
\newblock \emph{Neural Comput.}, 14\penalty0 (8):\penalty0 1771--1800, August
  2002.
\newblock ISSN 0899-7667.
\newblock \doi{10.1162/089976602760128018}.
\newblock URL \url{http://dx.doi.org/10.1162/089976602760128018}.

\bibitem[Hinton \& Salakhutdinov(2006)Hinton and
  Salakhutdinov]{hinton2006reducing}
Hinton, Geoffrey~E and Salakhutdinov, Ruslan~R.
\newblock Reducing the dimensionality of data with neural networks.
\newblock \emph{Science}, 313\penalty0 (5786):\penalty0 504--507, 2006.

\bibitem[Mooney \& Roy(2000)Mooney and Roy]{mooney2000content}
Mooney, Raymond~J and Roy, Loriene.
\newblock Content-based book recommending using learning for text
  categorization.
\newblock In \emph{Proceedings of the fifth ACM conference on Digital
  libraries}, pp.\  195--204. ACM, 2000.

\bibitem[Salakhutdinov et~al.(2007)Salakhutdinov, Mnih, and
  Hinton]{Salakhutdinov:2007:RBM:1273496.1273596}
Salakhutdinov, Ruslan, Mnih, Andriy, and Hinton, Geoffrey.
\newblock Restricted boltzmann machines for collaborative filtering.
\newblock In \emph{Proceedings of the 24th International Conference on Machine
  Learning}, ICML '07, pp.\  791--798, New York, NY, USA, 2007. ACM.
\newblock ISBN 978-1-59593-793-3.
\newblock \doi{10.1145/1273496.1273596}.
\newblock URL \url{http://doi.acm.org/10.1145/1273496.1273596}.

\bibitem[Symeonidis et~al.(2008)Symeonidis, Nanopoulos, and
  Manolopoulos]{symeonidis2008justified}
Symeonidis, Panagiotis, Nanopoulos, Alexandros, and Manolopoulos, Yannis.
\newblock Justified recommendations based on content and rating data.
\newblock In \emph{WebKDD Workshop on Web Mining and Web Usage Analysis}, 2008.

\bibitem[Tintarev \& Masthoff(2007)Tintarev and Masthoff]{tintarev2007survey}
Tintarev, Nava and Masthoff, Judith.
\newblock A survey of explanations in recommender systems.
\newblock In \emph{Data Engineering Workshop, 2007 IEEE 23rd International
  Conference on}, pp.\  801--810. IEEE, 2007.

\bibitem[Tintarev \& Masthoff(2011)Tintarev and
  Masthoff]{tintarev2011designing}
Tintarev, Nava and Masthoff, Judith.
\newblock Designing and evaluating explanations for recommender systems.
\newblock In \emph{Recommender Systems Handbook}, pp.\  479--510. Springer,
  2011.

\bibitem[Vig et~al.(2009)Vig, Sen, and Riedl]{vig2009tagsplanations}
Vig, Jesse, Sen, Shilad, and Riedl, John.
\newblock Tagsplanations: explaining recommendations using tags.
\newblock In \emph{Proceedings of the 14th international conference on
  Intelligent user interfaces}, pp.\  47--56. ACM, 2009.

\bibitem[Zanker(2012)]{Zanker:2012:IKE:2365952.2366011}
Zanker, Markus.
\newblock The influence of knowledgeable explanations on users' perception of a
  recommender system.
\newblock In \emph{Proceedings of the Sixth ACM Conference on Recommender
  Systems}, RecSys '12, pp.\  269--272, New York, NY, USA, 2012. ACM.
\newblock ISBN 978-1-4503-1270-7.
\newblock \doi{10.1145/2365952.2366011}.
\newblock URL \url{http://doi.acm.org/10.1145/2365952.2366011}.

\bibitem[Zhang(2015)]{zhang2015incorporating}
Zhang, Yongfeng.
\newblock Incorporating phrase-level sentiment analysis on textual reviews for
  personalized recommendation.
\newblock In \emph{Proceedings of the Eighth ACM International Conference on
  Web Search and Data Mining}, pp.\  435--440. ACM, 2015.

\bibitem[Zhang et~al.(2014)Zhang, Lai, Zhang, Zhang, Liu, and
  Ma]{zhang2014explicit}
Zhang, Yongfeng, Lai, Guokun, Zhang, Min, Zhang, Yi, Liu, Yiqun, and Ma,
  Shaoping.
\newblock Explicit factor models for explainable recommendation based on
  phrase-level sentiment analysis.
\newblock In \emph{Proceedings of the 37th international ACM SIGIR conference
  on Research \& development in information retrieval}, pp.\  83--92. ACM,
  2014.

\end{thebibliography}
\bibliographystyle{icml2012}

\end{document}